\documentclass[10pt,twocolumn,letterpaper]{article}

\usepackage[pagenumbers]{cvpr} 
\usepackage{multirow}
\usepackage[table]{xcolor} 
\definecolor{cvprblue}{rgb}{0.21,0.49,0.74}
\usepackage[pagebackref,breaklinks,colorlinks,allcolors=cvprblue]{hyperref}

\title{Decoupled Audio-Visual Dataset Distillation}

\author{Wenyuan Li \quad
Guang Li\thanks{Correspondence to: Guang Li (guang@lmd.ist.hokudai.ac.jp)} \quad
Keisuke Maeda \quad
Takahiro Ogawa \quad
Miki Haseyama \\ \\
Hokkaido University
}

\begin{document}
\maketitle
\begin{abstract}
Audio-Visual Dataset Distillation aims to compress large-scale datasets into compact subsets while preserving the performance of the original data. However, conventional Distribution Matching (DM) methods struggle to capture intrinsic cross-modal alignment. Subsequent studies have attempted to introduce cross-modal matching, but two major challenges remain: (i) independently and randomly initialized encoders lead to inconsistent modality mapping spaces, increasing training difficulty; and (ii) direct interactions between modalities tend to damage modality-specific (private) information, thereby degrading the quality of the distilled data. To address these challenges, we propose DAVDD, a pretraining-based decoupled audio–visual distillation framework. DAVDD leverages a diverse pretrained bank to obtain stable modality features and uses a lightweight decoupler bank to disentangle them into common and private representations. To effectively preserve cross-modal structure, we further introduce Common Intermodal Matching together with a Sample–Distribution Joint Alignment strategy, ensuring that shared representations are aligned both at the sample level and the global distribution level. Meanwhile, private representations are entirely isolated from cross-modal interaction, safeguarding modality-specific cues throughout distillation. Extensive experiments across multiple benchmarks show that DAVDD achieves state-of-the-art results under all IPC settings, demonstrating the effectiveness of decoupled representation learning for high-quality audio–visual dataset distillation. Code will be released.
\end{abstract}    
\section{Introduction}
\label{sec:intro}

Recent advances in deep learning have enabled Deep Neural Networks (DNNs) to deliver substantial performance gains across a wide range of tasks~\cite{Gaurav2023DLsurvey}. However, these improvements have been accompanied by a rapid increase in data scale, resulting in prohibitive storage demands and rising computational costs~\cite{Mustafa2023DLsurvey}. Dataset Distillation (DD) addresses this issue by synthesizing a compact set of samples that preserves the essential information of the original dataset, thereby reducing data size while maintaining model performance~\cite{wang2018datasetdistillation, li2022awesome}. Over the years, DD research has expanded into multiple branches, including trajectory matching~\cite{MTT, t2, li2023ddpp, li2024iadd}, gradient matching~\cite{DC, DSA}, distribution matching~\cite{DM, DM1, DM2, li2025hdd}, and generative apporach~\cite{gu2024efficient, li2024generative, li2025diffusion, li2025diff, ye2025igds}, each demonstrating strong potential in applications such as neural architecture search~\cite{n2, n3}, continual learning~\cite{c1, c2, sangermano2022sample}, and privacy-sensitive learning~\cite{p1, p2, li2022compressed, li2023sharing}.

\begin{figure}[t]
\centerline{\includegraphics[width=1\linewidth]{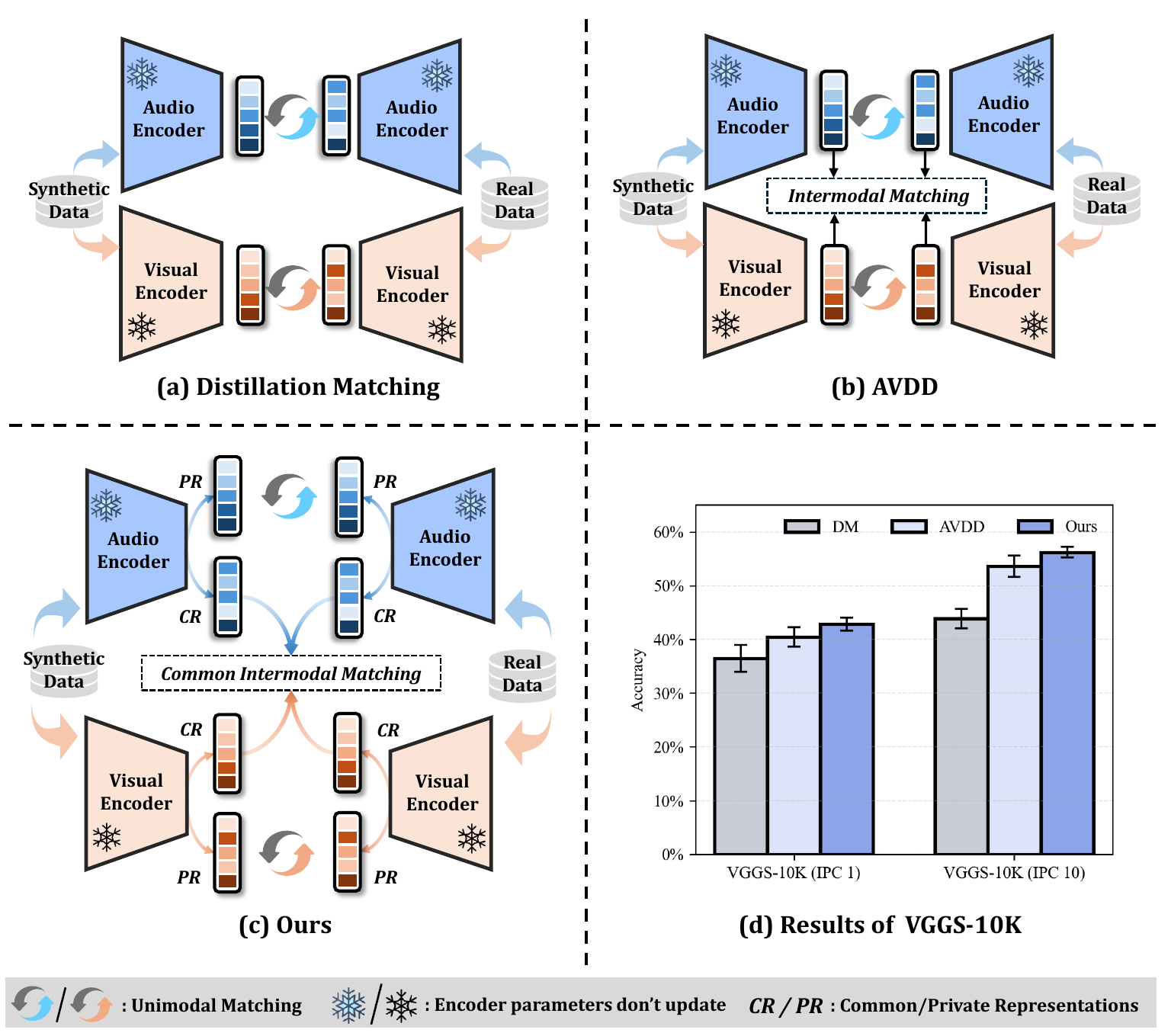}}
\caption{A comparison of architectures for multimodal audio-visual dataset distillation. Subfigures (a), (b), and (c) depict the frameworks of Distillation Matching (DM), Audio-Visual Dataset Distillation (AVDD), and the proposed Decoupled Audio-Visual Dataset Distillation (DAVDD), respectively. Subfigure (d) reports performance on the VGGS-10K dataset under IPC 1 and IPC 10 settings, where DAVDD consistently surpasses prior methods in both accuracy and variance.}
\label{overview}
\end{figure}

Among existing approaches, Distribution Matching (DM) has gained particular popularity due to its strong empirical performance and efficient optimization procedure~\cite{DM}. DM seeks to align the feature-space distributions of synthetic samples with those of real data, enabling models trained on the distilled dataset to achieve accuracy comparable to training on the full dataset. However, DM has been most successful in single-modal settings such as image classification, where alignment is limited to intra-modal distributions and cross-modal structure is not considered~\cite{DM1}.

With the rise of multimodal audio-visual learning~\cite{imagebind,cAV-MAE,AV-encoder}, dataset sizes have expanded even further, yet most existing distillation methods remain confined to single-modality settings. As a result, distillation for audio-visual datasets is still underexplored, and a central challenge persists: how to preserve inter-modal correlations during synthesis. Conventional DM-based methods operate solely through independent within-modality matching, which maintains modality-specific information but entirely overlooks cross-modal relationships (see Fig.~\ref{overview}(a)). To mitigate this limitation, AVDD introduces a cross-modal matching loss that aligns synthetic samples in a joint feature space, enabling the distilled set to retain audio-visual associations better (see Fig.~\ref{overview}(b))~\cite{AVDD}.

However, the joint optimization strategy used in AVDD still faces two fundamental limitations. First, it depends on independently initialized encoders with random parameters, causing the visual and audio branches to operate in mismatched embedding spaces. This modality gap significantly complicates cross-modal interaction. Second, enforcing cross-modal matching directly during distillation forces the synthetic samples to satisfy both intra-modal and inter-modal objectives simultaneously, which can introduce conflicting constraints. As a result, modality-private information is often distorted or lost, ultimately reducing the fidelity of the distilled dataset.

To address these limitations, we propose DAVDD, a pretraining-based and fully decoupled audio–visual dataset distillation framework (see Fig.~\ref{overview}(c)). We first pretrain multiple audio and visual encoder pairs, each trained independently, and freeze their weights to construct a diverse pretrained model bank. For every encoder pair, we attach a lightweight decoupler bank composed of two-layer MLPs that project encoder outputs into a shared space. These decouplers disentangle the features into common representations for cross-modal reasoning and private representations, where the original encoder output is preserved to retain modality-specific information.

To enhance shared-feature alignment, we introduce a \textit{Sample–Distribution Joint Alignment strategy} that combines intra-sample and inter-sample contrastive learning with distribution-level matching. During distillation, we perform intra-modal matching exclusively on private representations, preventing cross-modal interference and safeguarding modality-specific cues. In parallel, we apply \textit{Common Intermodal Matching} on the common representations to robustly capture cross-modal correlations within the synthetic dataset. Experiments across multiple benchmarks confirm that DAVDD achieves state-of-the-art performance.

To summarize, our contributions are as follows:

\begin{itemize}
    \item We propose a novel pretraining-based and decoupled framework for audio-visual dataset distillation that jointly improves efficiency while preserving both modality-private information and cross-modal correlations.
    \item We design a Common Intermodal Matching mechanism that enforces cross-modal alignment while maintaining the integrity of modality-specific private features.
    \item We introduce a Sample–Distribution Joint Alignment strategy that combines intra-sample and inter-sample contrastive learning with distribution-level matching to substantially enhance the consistency of shared representations.
    \item We conduct extensive experiments across multiple benchmarks, demonstrating consistent and substantial performance gains over previous state-of-the-art distillation methods.
\end{itemize}

\section{Related Works}
\label{related}

\subsection{Audio-Visual Learning}
Video inherently contains synchronized and complementary audio-visual signals. Leveraging this natural cross-modal consistency, recent work has advanced audio-visual learning across several fronts, including self-supervised representation learning and cross-modal alignment~\cite{imagebind,cAV-MAE,AV-encoder}, sound source separation and localization~\cite{iquery,AV-seg,AV-group}, and event or action recognition with acoustic guidance~\cite{open-local,AV-local}. These studies collectively show that audio-visual coherence provides both a powerful supervisory signal and a strong inductive bias, enabling the learning of transferable and alignment-aware multimodal representations that consistently outperform unimodal counterparts. In this work, we extend these insights to the dataset distillation setting, aiming to enhance the quality of synthesized audio-visual data for audio-visual event recognition~\cite{2018audio}.

\subsection{Shared-Private Representation Learning}
Decomposing multimodal or multi-view data into shared and private representations has proven effective for enhancing downstream performance and is widely adopted across various domains, including multi-view graph learning~\cite{R38,R39,R40}, multimodal sentiment analysis~\cite{R32,R33,R35,R37}, multimodal remote-sensing recognition~\cite{R41,R36,R42,R43}, and related areas~\cite{R31,R312}. Inspired by this principle, we extend shared–private modeling to multimodal dataset distillation, to capture cross-modal correlations while preserving modality-specific information, thereby avoiding the inherent trade-off between shared and private representations.

\begin{figure*}[ht]    \centerline{\includegraphics[width=1\linewidth]{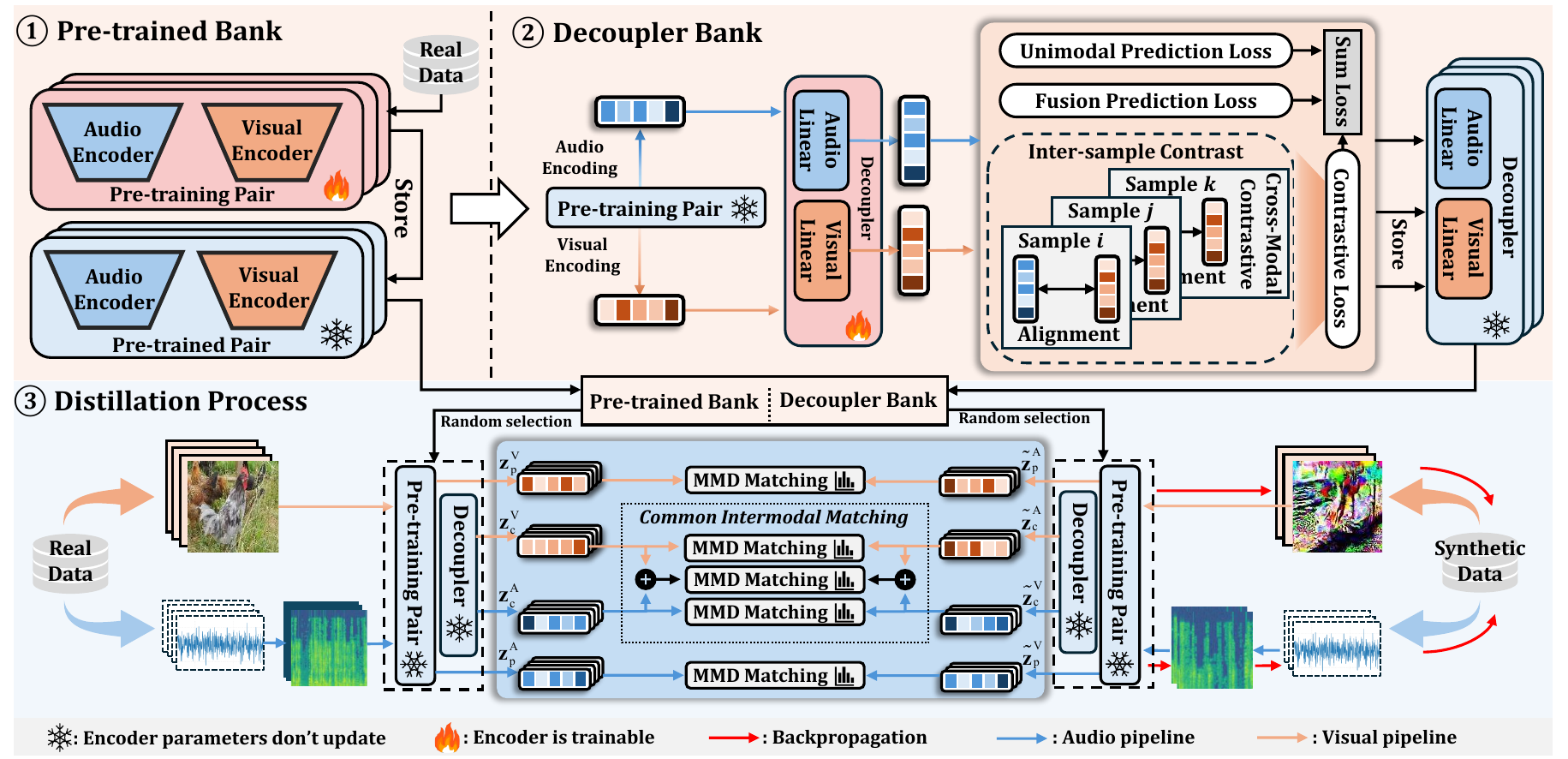}}
\caption{Overview of the DAVDD framework. DAVDD consists of three key components: a pre-trained model bank providing stable and diverse audio–visual encoders, a decoupler bank that projects encoder outputs into shared and private representation spaces, and a decoupled distillation process that performs intra-modal matching on private features and cross-modal alignment on shared features. By combining Sample–Distribution Joint Alignment with Common Intermodal Matching, DAVDD preserves modality-specific information while effectively capturing audio–visual correlations, enabling the synthesis of high-fidelity multimodal datasets.}
\label{main-figure}
\end{figure*}

\section{Methodology}

\subsection{Problem Setting}
For the $i$-th training sample, let $\mathbf{a}_i$ and $\mathbf{v}_i$ denote the audio waveform and the video frame, respectively, and define the paired input as $\mathbf{z}_i=(\mathbf{a}_i,\mathbf{v}_i)$ with label $y_i$.
Given a large audio–visual dataset
$\mathcal{D}=\{(\mathbf{z}_i,y_i)\}_{i=1}^{|\mathcal{R}|}$, the objective of audio–visual dataset distillation is to construct a much smaller \emph{synthetic set}
$\widetilde{\mathcal{D}}=\{(\widetilde{\mathbf{z}}_j,\widetilde{y}_j)\}_{j=1}^{|\mathcal{S}|}$,
with ${|\mathcal{S}|}\!\ll\! {|\mathcal{R}|}$, where each synthetic sample $\widetilde{\mathbf{z}}_j=(\widetilde{\mathbf{a}}_j,\widetilde{\mathbf{v}}_j)$ contains both audio and visual components.
The distilled set should retain the essential information in $\mathcal{D}$ such that a model trained solely on $\widetilde{\mathcal{D}}$ achieves performance comparable to training on the full dataset.
Formally, let $\phi_\theta$ denote a neural network trained on a dataset and $\ell(\cdot,\cdot)$ a loss function such as cross-entropy.
Denote by $\theta^\star(\mathcal{D})$ and $\theta^\star(\widetilde{\mathcal{D}})$ the parameters obtained after training $\phi_\theta$ on $\mathcal{D}$ and $\widetilde{\mathcal{D}}$, respectively. The goal of distillation is to ensure:
\begin{equation}
\begin{aligned}
&\mathbb{E}_{(z,y)\sim \mathcal{D}_{\mathrm{test}}}\!\left[\ell\big(\phi_{\theta^\star(\mathcal{D})}(z),y\big)\right] \\
&\;\approx\;
\mathbb{E}_{(z,y)\sim \mathcal{D}_{\mathrm{test}}}\!\left[\ell\big(\phi_{\theta^\star(\widetilde{\mathcal{D}})}(z),y\big)\right].
\end{aligned}
\label{eq:1}
\end{equation}
To operationalize Eq.~(\ref{eq:1}), optimization-driven approaches in dataset distillation typically fall into two camps: 
(i) meta-learning updates that directly refine the synthetic set $\widetilde{\mathcal{D}}$; and 
(ii) gradient/parameter matching between models trained on $\widetilde{\mathcal{D}}$ and on the real data $\mathcal{D}$.
Both are formulated as bi-level programs and thus require differentiating through an inner training loop, which is computationally expensive.

In contrast, DM~\cite{DM} avoids nested gradients by matching the \emph{representation distributions} induced by random feature extractors on $\mathcal{D}$ and $\widetilde{\mathcal{D}}$. 
Within this paradigm, alignment can be conducted at the instance level or via moment matching.
Instance-wise alignment captures local correspondences but may fail to reflect global corpus-level structure; accordingly, we adopt moment matching, formulated as:
\begin{equation}
\begin{split}
\widetilde{\mathcal{D}}^{\star}
&= \arg\min_{\widetilde{\mathcal{D}}}\; \mathbb{E}_{\varphi \sim \mathcal{P}}
\Bigg\|
\frac{1}{|\mathcal{D}|}\sum_{(\mathbf{z},y)\in\mathcal{D}}\varphi(\mathbf{z})
\\[-2pt]
&\qquad-
\frac{1}{|\widetilde{\mathcal{D}}|}\sum_{(\tilde{\mathbf{z}},\tilde{y})\in\widetilde{\mathcal{D}}}\varphi(\tilde{\mathbf{z}})
\Bigg\|^{2},
\end{split}
\end{equation}
where $\varphi\!\sim\!\mathcal{P}$ denotes a feature extractor sampled from a distribution (e.g., a randomly initialized deep network without the final linear classification head).

To assess audio--visual dataset distillation~\cite{AVDD}, we use an audio--visual event recognition proxy.
Each sample has audio $\mathbf{a}$ and video $\mathbf{v}$. An audio encoder $\psi^{a}$ processes the spectrogram $S(\mathbf{a})$ to obtain an audio feature $f^{a}$, while a visual encoder $\psi^{v}$ maps $\mathbf{v}$ to a visual feature $f^{v}$. A fusion operator $\mathcal{F}$ combines them into a joint representation $u=\mathcal{F}(f^{a},f^{v})$, which is then passed to a classifier to produce prediction probabilities $\mathbf{p}$. Training follows the standard cross-entropy objective:
\begin{equation}
\mathcal{L}_{\mathrm{CE}}(y,\mathbf{p})=-\sum_{c=1}^{|\mathcal{C}|} y_c \log p_c .
\end{equation}
This proxy assesses whether the synthetic audio–visual data effectively supports multimodal recognition and how different fusion strategies influence performance.

\subsection{Overview}
In this paper, we propose a pretrained Decoupled Audio–Visual Dataset Distillation (DAVDD) framework.
DAVDD performs fine-grained matching in both the common and private representation spaces, enabling the synthesis of a compact audio–visual dataset whose feature distribution closely aligns with that of the original data.
An overview of the full architecture is provided in Fig.~\ref{main-figure}. The key idea is to leverage multiple high-quality pretrained feature extractors and, for each, attach a lightweight decoupler bank that separates encoder outputs into modality-private and cross-modality-common components. This design effectively mitigates the challenges posed by direct cross-modal interaction during matching and supports stable, decoupled optimization across feature spaces.

\subsubsection{Pre-trained Bank}
We begin by constructing a pre-trained bank composed of multiple audio–visual encoder pairs, each offering a distinct feature perspective for subsequent distillation.
This diversity enables multi-view alignment during matching and improves the robustness of the distilled representations.
Concretely, we independently train $M$ audio encoders and visual encoders on $\mathcal{D}$, and store their learned weights as reusable feature extractors. For the $m$-th pre-trained encoder pair, given an input $(\mathbf{a}_i,\mathbf{v}_i)$, the feature extraction process is defined as:
\begin{equation}
\varphi^{m}
= \big(\psi^{a}_{m}(S(\mathbf{a}_i);\,\theta^{a\dagger}_{m}),\;
       \psi^{v}_{m}(\mathbf{v}_i;\,\theta^{v\dagger}_{m})\big),
\label{eq:pair-def}
\end{equation}
where, for $m=1, \ldots, M$, $\theta^{a\dagger}_{m}$ and $\theta^{v\dagger}_{m}$ denote the parameters obtained from the independent fits on $\mathcal{D}$. The pre-trained bank can then be represented as:
\begin{equation}
\mathcal{B}_{\mathrm{pre}}
=\big\{\varphi^{m}\big\}_{m=1}^{M}.
\label{eq:bank-def}
\end{equation}
Entries retrieved from $\mathcal{B}_{\mathrm{pre}}$ are kept frozen in downstream stages.

\subsubsection{Decoupler Bank}
For each $\varphi^{m}\!\in\!\mathcal{B}_{\mathrm{pre}}$, a dedicated decoupler bank is attached:
\begin{equation}
\label{eq:dec-lib}
\begin{aligned}
\mathcal{B}^{m}_{\mathrm{dec}} &= \big\{\delta^{(m,t)}\big\}_{t=1}^{T},\\
\delta^{(m,t)} &= \big(g^{a}_{m,t},\, g^{v}_{m,t}\big),
\end{aligned}
\end{equation}
where $g^{a}_{m,t}$ and $g^{v}_{m,t}$ are lightweight linear maps into a shared space $\mathbb{R}^{d_c}$.
We collect all decoupler banks as:
\begin{equation}
\mathcal{B}_{\mathrm{dec}}=\big\{\mathcal{B}^{m}_{\mathrm{dec}}\big\}_{m=1}^{M}.
\label{eq:all-dec}
\end{equation}

Given an input $(\mathbf{a}_i,\mathbf{v}_i)$ and the frozen encoders in $\varphi^{(m)}$, the decoupler projections can be represented as:
\begin{equation}
\begin{aligned}
c^{a}_{i,m,t}=g^{a}_{m,t}(\theta^{a\dagger}_{m}),\\
c^{v}_{i,m,t}=g^{v}_{m,t}(\theta^{v\dagger}_{m}).
\end{aligned}
\label{eq:proj}
\end{equation}

The decouplers are optimized with a joint loss to (i) broaden multi-view coverage across $t=0,\ldots,T$ and
(ii) promote a common representation; the objective is deferred to Subsection 3.3.

\subsubsection{Distillation Process}
Given the original dataset $ \mathcal{D}$ and the synthetic dataset for update $ \widetilde{\mathcal{D}}$, 
we first uniformly and randomly sample a pre-trained pair $\varphi^{m}$ from the $\mathcal{B}_{\mathrm{pre}}$ for each batch of real and synthetic data, and subsequently sample $\delta^{(m,t)}$ from the $\mathcal{B}^{m}_{\mathrm{dec}}$ to $\varphi^{m}$.

Subsequently, leveraging the sampled \( \varphi^{m} \) and \( \delta^{(m,t)} \), we encode the data. We reinterpret the direct output of \( \varphi^{m} \) as the private representation, while the output obtained via a secondary encoding through \( \delta^{(m,t)} \) is regarded as the modality-common representation, as follows:
\begin{equation}
\label{eq:dec-lib2}
\begin{aligned}
\mathbf{z}_{p} \;&=\; \varphi^{m}\!(\cdot), \\
\mathbf{z}_{c} \;&=\; \delta^{(m,t)}\!(\mathbf{z}_{p} ).
\end{aligned}
\end{equation}

Then, for the $i$-th sample $\mathbf{z}_i = (\mathbf{a}_i,\mathbf{v}_i)$  in the real data, the common representation and private representation can be expressed as:
\begin{equation}
\begin{aligned}
\mathbf{z}^{A}_{p}(i) &= \varphi^{m}\!(S(\mathbf{a}_i)),\  \mathbf{z}^{A}_{c}(i) = \delta^{(m,t)}\!(\mathbf{z}^{A}_{p}(i)),\\
\mathbf{z}^{V}_{p}(i) &= \varphi^{m}\!(\mathbf{v}_i),\  \mathbf{z}^{V}_{c}(i) = \delta^{(m,t)}\!(\mathbf{z}^{V}_{p}(i)),
\end{aligned}
\end{equation}
where $\mathbf{z}^{A}_{p}(i)$ and $\mathbf{z}^{A}_{c}(i)$ denote the private and common representations of the audio modality for the $i$-th sample, respectively, and $\mathbf{z}^{V}_{p}(i)$ and $\mathbf{z}^{V}_{c}(i)$ denote the private and common representations of the visual modality for the $i$-th sample, respectively.

Similarly, for the $j$-th sample of the synthetic data $\tilde{\mathbf{z}}_j = (\tilde{\mathbf{a}}_j, \tilde{\mathbf{v}}_j)$, the common and private representations can be formulated as follows:
\begin{equation}
\begin{aligned}
\mathbf{\tilde{z}}^{A}_{p}(j) &= \varphi^{m}\!(S(\tilde{\mathbf{a}}_j)), \, \mathbf{\tilde{z}}^{A}_{c}(j) = \delta^{(m,t)}\!(\mathbf{\tilde{z}}^{A}_{p}(j)),\\
\mathbf{\tilde{z}}^{V}_{p}(j) &= \varphi^{m}\!(\tilde{\mathbf{v}}_j), \, \mathbf{\tilde{z}}^{V}_{c}(j) = \delta^{(m,t)}\!(\mathbf{\tilde{z}}^{V}_{p}(j)),
\end{aligned}
\end{equation}
where $\mathbf{\tilde{z}}^{A}_{p}(j)$ and $\mathbf{\tilde{z}}^{A}_{c}(j)$ denote the private and common representations of the audio modality for the $j$-th synthetic sample, respectively, and $\mathbf{\tilde{z}}^{V}_{p}(j)$ and $\mathbf{\tilde{z}}^{V}_{c}(j)$ denote the private and common representation of the visual modality for the $j$-th synthetic sample.

\textbf{Private Representation Matching} For the private representation branch, DAVDD follows the unimodal structure of DM and does not introduce any cross-modal interaction.
Instead, it independently matches the audio and visual private representations of the real and synthetic data.
This design ensures that modality-specific cues are preserved without being influenced by cross-modal constraints, thereby maintaining the full fidelity of private information encoded in the original dataset.
The Private Representation Matching loss $\mathcal{L}_{\text{pr}}$ is formulated as:
\begin{equation}
\mathcal{L}_{\text{pr}}
= \mathcal{L}^{A}_{\text{pr}}+\mathcal{L}^{V}_{\text{pr}},
\end{equation}
where $\mathcal{L}^{A}_{\text{pr}}$ and $\mathcal{L}^{V}_{\text{pr}}$ denote:
\begin{equation}
\mathcal{L}^{A}_{\text{pr}}
= \left\lVert
\frac{1}{|\mathcal{R}|} \sum_{i=1}^{|\mathcal{R}|}
\mathbf{z}^{A}_{p}(i)
-
\frac{1}{|\mathcal{S}|} \sum_{j=1}^{|\mathcal{S}|}
\mathbf{\tilde{z}}^{A}_{p}(j)
\right\rVert^{2},
\end{equation}
\begin{equation}
\mathcal{L}^{V}_{\text{pr}}
= \left\lVert
\frac{1}{|\mathcal{R}|} \sum_{i=1}^{|\mathcal{R}|}
\mathbf{z}^{V}_{p}(i)
-
\frac{1}{|\mathcal{S}|} \sum_{j=1}^{|\mathcal{S}|}
\mathbf{\tilde{z}}^{V}_{p}(j)
\right\rVert^{2}.
\end{equation}

Moreover, prior studies suggest that shared and private representations may exhibit non-trivial dependencies~\cite{time-series}.
By matching private representations independently within each modality, DAVDD preserves these underlying relationships without disrupting modality-specific structure, leading to more robust and faithful synthetic data.

\textbf{Common Intermodal Matching} For the shared-feature branch, we first align the common representations extracted from each unimodal pathway to ensure that both audio and visual modalities in the synthetic data faithfully capture their corresponding cross-modal shared structure.
Formally, the common representations are defined as follows:
\begin{equation}
\mathcal{L}^{A}_{\text{com}}
= \left\lVert
\frac{1}{|\mathcal{R}|} \sum_{i=1}^{|\mathcal{R}|}
\mathbf{z}^{A}_{c}(i)
-
\frac{1}{|\mathcal{S}|} \sum_{j=1}^{|\mathcal{S}|}
\mathbf{\tilde{z}}^{A}_{c}(j)
\right\rVert^{2},
\end{equation}

\begin{equation}
\mathcal{L}^{V}_{\text{com}}
= \left\lVert
\frac{1}{|\mathcal{R}|} \sum_{i=1}^{|\mathcal{R}|}
\mathbf{z}^{V}_{c}(i)
-
\frac{1}{|\mathcal{S}|} \sum_{j=1}^{|\mathcal{S}|}
\mathbf{\tilde{z}}^{V}_{c}(j)
\right\rVert^{2}.
\end{equation}

During the distillation process, the synthetic samples evolve continuously, and the correlation between their audio and visual modalities changes accordingly.
To maintain stable cross-modal alignment throughout this optimization, we introduce a modality-shared joint matching objective as an additional consistency constraint.
This serves as a secondary mechanism that reinforces the preservation of audio–visual relationships, formulated as follows:
\begin{equation}
\begin{aligned}
\bar{R}^{AV}_{com}
= \biggl[
\frac{1}{|\mathcal{R}|} \sum_{i=1}^{|\mathcal{R}|}
\mathbf{z}^{A}_{c}(i)
+
\frac{1}{|\mathcal{R}|} \sum_{i=1}^{|\mathcal{R}|}
\mathbf{z}^{V}_{c}(i)
\biggr], \\[0.5em]
\bar{S}^{AV}_{com}
= \biggl[
\frac{1}{|\mathcal{S}|} \sum_{j=1}^{|\mathcal{S}|}
\mathbf{\tilde{z}}^{A}_{c}(j)
+
\frac{1}{|\mathcal{S}|} \sum_{j=1}^{|\mathcal{S}|}
\mathbf{\tilde{z}}^{V}_{c}(j)
\biggr], 
\end{aligned}
\end{equation}

\begin{equation}
\begin{aligned}
\mathcal{L}^{AV}_{com} &= \left\lVert \bar{R}^{AV}_{com} - \bar{S}^{AV}_{com} \right\rVert^{2}.
\end{aligned}
\end{equation}

Therefore, our Common Intermodal Matching loss $\mathcal{L}_{com}$ is defined as:
\begin{equation}
\begin{aligned}
\mathcal{L}_{com} &= \mathcal{L}^{AV}_{com} + \mathcal{L}^{A}_{com} + \mathcal{L}^{V}_{com}.
\end{aligned}
\end{equation}

Furthermore, the overall objective of DAVDD combines all components described above.
The total distillation loss $\mathcal{L}$ is therefore defined as:
\begin{equation}
\begin{aligned}
\mathcal{L}_{dis} &= \lambda_{c}\mathcal{L}_{com}  + \lambda_{p}\mathcal{L}_{pr},
\end{aligned}
\end{equation}
where $\lambda_{\text{c}}$ and $\lambda_{\text{p}}$ are the corresponding weighting factors.
The total distillation loss explicitly decouples private and common representation learning, enabling DAVDD to simultaneously preserve modality-specific information and enforce stable cross-modal alignment.

\subsection{Decoupling Process}

To ensure the spatial or semantic alignment of the shared feature representations between the audio and visual modalities output by the decoupler, while maintaining the semantic validity of these shared components, we design a decoupling process with a joint loss function to impose constraints on them.

Specifically, we first apply a classification loss to the common representations in each modality, guiding the model to learn discriminative shared semantic information. Additionally, a fusion loss is introduced to enhance the synergy and consistency of the shared information across different modalities, thereby ensuring that the semantic information carried by the shared components remains meaningful. The details are as follows:
\begin{equation}
\begin{aligned}
\mathcal{L}_{\text{cls}} = & \lambda_{\text{com}} \left( \mathcal{L}_{\text{CE}}(l_{a}, y) + \mathcal{L}_{\text{CE}}(l_{v}, y) \right) \\
& + \lambda_{\text{fu}} \mathcal{L}_{\text{CE}}(l_{\text{fuse}}, y).
\end{aligned}
\end{equation}
Here, $\mathcal{L}_{\text{CE}}$ denotes the Cross-Entropy Loss. The terms $l_{a}$, $l_{v}$, and $l_{\text{fuse}}$ represent the classification logits derived respectively from the audio features, visual features, and the fused features (where the two modalities are concatenated along the feature dimension). Additionally, $\lambda_{\text{com}}$ and $\lambda_{\text{fu}}$ are the weighting parameters corresponding to each component.

Furthermore, we observe that the real dataset exhibits significant intra-class variations and a dispersed shared semantic space, which hinders the semantic representation of the encoded generated dataset~\cite{IID}. To address this, we introduce inter-sample contrastive learning to construct a unified category-level semantic space. First, define the composite features $z_{i}^{\text{inter}}$ utilized for this loss:
\begin{equation}
\begin{aligned}
z_{i}^{\text{inter}} 
= \mathcal{N}\bigl( \mathbf{z}^{A}_{c}(i) + \mathbf{z}^{V}_{c}(i) \bigr),
\end{aligned}
\end{equation}
where $\mathcal{N}(x) = {x}/{\|x\|_2}$, denoting the $L2$ normalization operation. Therefore, our inter-sample contrastive learning loss is formulated as:
\begin{equation}
\begin{aligned}
&\mathcal{L}_{\mathrm{inter}}
= \sum_{i \in I} \frac{-1}{|W(i)|} \\&\sum_{w \in W(i)}
\log
\frac{
    \exp\big(\mathrm{sim}(z_i^{\text{inter}}, z_w^{\text{inter}})/\tau\big)
}{
    \sum_{k \in I \setminus \{i\}}
    \exp\big(\mathrm{sim}(z_i^{\text{inter}},z_k^{\text{inter}})/\tau\big),
}
\end{aligned}
\end{equation}
where $W(i) = \{w \in I \setminus \{i\} \mid y_w = y_i\}$ denotes the set of positive samples within the batch that share the same class label as sample $i$ (excluding $i$ itself). $\tau$ is the temperature hyperparameter.

\textbf{Sample–Distribution Joint Alignment} To achieve more comprehensive alignment between the two modalities and to fully learn the modality correlations, we propose a Sample–Distribution Joint Alignment strategy. This approach not only performs cross-modal alignment at the sample level but also aligns modalities across the overall dataset distribution. Specifically, for sample-level cross-modal alignment, following prior methodologies~\cite{clip}, we treat the alternative modality of the same sample as a positive sample, and the alternative modalities of all other samples within the same batch as negative samples. 
Given a minibatch of $N$, the audio-to-visual cross-modal contrastive loss is defined as:
\begin{equation}
\mathcal{L}_{\mathrm{A}\to\mathrm{V}}
= -\frac{1}{N} \sum_{i=1}^{N}
\log
\frac{
    \exp\!\left( \operatorname{sim}\big(\mathbf{z}^{A}_{c}(i), \mathbf{z}^{V}_{c}(i)\big) / \tau \right)
}{
    \sum_{j=1}^{N}
    \exp\!\left( \operatorname{sim}\big(\mathbf{z}^{A}_{c}(i), \mathbf{z}^{V}_{c}(j)\big) / \tau \right)
}.
\label{eq:intra_a2v}
\end{equation}
Similarly, the visual-to-audio cross-modal contrastive loss is:
\begin{equation}
\mathcal{L}_{\mathrm{V}\to\mathrm{A}}
= -\frac{1}{N} \sum_{i=1}^{N}
\log
\frac{
    \exp\!\left( \operatorname{sim}\big(\mathbf{z}^{V}_{c}(i), \mathbf{z}^{A}_{c}(i)\big) / \tau \right)
}{
    \sum_{j=1}^{N}
    \exp\!\left( \operatorname{sim}\big(\mathbf{z}^{V}_{c}(i), \mathbf{z}^{A}_{c}(j)\big) / \tau \right)
}.
\label{eq:intra_v2a}
\end{equation}
The overall sample-level cross-modal alignment loss is defined as the symmetric average:
\begin{equation}
\mathcal{L}_{\mathrm{intra}}
= \frac{1}{2}
\left(
    \mathcal{L}_{\mathrm{A}\to\mathrm{V}}
    + \mathcal{L}_{\mathrm{V}\to\mathrm{A}}
\right).
\label{eq:intra_total}
\end{equation}

Both inter-sample and intra-sample contrastive learning are computed within a mini-batch, and their perspective is therefore constrained by the batch size. To achieve a more global and stable alignment objective, we introduce a distribution-level alignment loss. The core idea of this loss is to align the mean feature of each class in the current batch ($\mu$) with the historically observed cross-modal global class prototypes ($P$).

For each class $c$, we maintain exponential moving average (EMA) prototypes for the audio and visual modalities, denoted by $P^{A}_{c}$ and $P^{V}_{c}$, respectively. At each training step, we first compute the mean features $\mu^{A}_{c}$ and $\mu^{V}_{c}$ for the set of classes $\mathcal{C}_{\mathcal{B}}$ that appear in the current batch $\mathcal{B}$. The alignment loss is then defined as the cosine distance between the current means and the cross-modal global prototypes, as follows:
\begin{equation}
\mathcal{L}_{\text{align}} = \frac{1}{|\mathcal{C}_{\mathcal{B}}|} 
\sum_{c \in \mathcal{C}_{\mathcal{B}}} \left[ 
d_{\cos}(\mu^{A}_{c}, P^{V}_{c}) + d_{\cos}(\mu^{V}_{c}, P^{A}_{c}) 
\right].
\end{equation}
Here, $d_{\cos}(u, v) = 1 - u^\top v$. $P^{A}_{c}$ and $P^{V}_{c}$ are treated as fixed targets and are excluded from gradient back-propagation. The EMA prototype bank $P$ is updated asynchronously after the optimizer step (see Appendix A). This mechanism drives the batch-level feature means toward a stable, cross-modal global distribution centroid, thereby substantially enhancing the robustness of the alignment.

Therefore, our overall decoupling loss is:
\begin{equation}
\mathcal{L}_{\text{de}} = \mathcal{L}_{\text{cls}} + \lambda_{\text{inter}} \mathcal{L}_{\text{inter}} + \lambda_{\text{intra}} \mathcal{L}_{\text{intra}} + \lambda_{\text{align}} \mathcal{L}_{\text{align}},
\end{equation}
where $\lambda_{\text{inter}}$, $\lambda_{\text{intra}}$, and $\lambda_{\text{align}}$ are the corresponding weighting factors.
The decoupling loss provides a unified supervisory signal that enforces discriminative shared semantics, stabilizes cross-modal alignment, and promotes consistent audio–visual structure across both local batches and global feature distributions.

\section{Experiments}

\begin{table*}[t]
\centering
\caption{Comparison of different methods on the VGGS-10K, MUSIC-21, and AVE datasets with IPC = 1, 10, and 20.}
\label{main}
\scalebox{0.85}{
\begin{tabular}{lccccccccc}
\toprule
\multirow{1}{*}{Method} 
& \multicolumn{3}{c}{VGGS-10K} 
& \multicolumn{3}{c}{MUSIC-21} 
& \multicolumn{3}{c}{AVE} \\
\cmidrule{1-10}
IPC & 1 & 10 & 20 & 1 & 10 & 20 & 1 & 10 & 20 \\
Ratio (\%) & 0.11 &0.17& 0.83 & 0.02 &0.2& 1 & 0.2 & 
2 & 10 \\
\midrule
Random     & 15.4$\pm$1.9 & 32.0$\pm$1.6 & 45.1$\pm$2.3 & 24.1$\pm$4.3 & 45.8$\pm$1.7 & 54.9$\pm$1.9 & 10.0$\pm$1.2 & 20.0$\pm$1.5 & 26.3$\pm$1.0 \\
Herding~\cite{herd}    & 20.8$\pm$2.1 & 39.9$\pm$1.6 & 50.2$\pm$0.7 & 26.2$\pm$2.0 & 51.9$\pm$1.4 & 60.0$\pm$0.9 & 11.8$\pm$0.4 & 26.9$\pm$0.5 & 33.0$\pm$0.4 \\
\midrule
DC~\cite{DC}   & 18.3$\pm$1.4 & 32.1$\pm$0.8 & - & 22.6$\pm$1.1 & - & - & 10.5$\pm$0.4 & - & -\\
DSA~\cite{DSA}        & 19.3$\pm$1.4 & 36.6$\pm$1.0 & - & 23.0$\pm$1.2 & - & - & 10.8$\pm$0.6 & - & - \\
MTT~\cite{MTT}     & 34.1$\pm$3.6 & 36.8$\pm$2.0 & 51.9$\pm$1.3 & 28.7$\pm$1.2 & 42.3$\pm$1.1 & - & 12.1$\pm$0.4 & 23.2$\pm$1.0 & - \\
\midrule
DM~\cite{DM}         & 36.5$\pm$2.5 & 43.9$\pm$1.8 & 49.0$\pm$2.4 & 38.3$\pm$1.3 & 54.8$\pm$1.4 & 61.1$\pm$1.3 & 21.7$\pm$1.5 & 28.1$\pm$1.8 & 32.6$\pm$1.0 \\
AVDD~\cite{AVDD}        & 40.2$\pm$1.6 & 54.0$\pm$1.8 & 57.7$\pm$1.7 & 44.2$\pm$2.0 & 66.9$\pm$1.2 & 69.7$\pm$0.8 & 22.7$\pm$1.5 & 35.9$\pm$0.9 & 40.0$\pm$1.2 \\
 \rowcolor{gray!30}  \textbf{DAVDD}   & \textbf{41.9$\pm$1.2} & \textbf{56.2$\pm$1.0} & \textbf{59.1$\pm$1.2} & \textbf{45.7$\pm$1.8} & \textbf{67.8$\pm$0.9} & \textbf{71.4$\pm$0.9} & \textbf{23.6$\pm$1.1} & \textbf{37.3$\pm$0.6} & \textbf{40.9$\pm$0.7} \\
\midrule
Whole Dataset & &68.2$\pm$0.8&  &  &85.9$\pm$0.8&  &  &52.2$\pm$0.4&  \\
\bottomrule
\end{tabular}}
\end{table*}

\subsection{Experimental setup}

\subsubsection{Datasets}

We adhere to the dataset processing approach and experimental settings employed by AVDD~\cite{AVDD} and accordingly evaluate our method on the three datasets.

\textbf{VGGS-10K.} AVDD~\cite{AVDD} curated a subset from VGGSound~\cite{vggsound} named VGGS-10K. This subset randomly samples 10 categories and contains 8,808 training videos and 444 test videos. VGGSound is a large audio-visual dataset comprising approximately 200,000 video clips sourced from YouTube, spanning 309 distinct categories.

\textbf{MUSIC-21.} MUSIC-21~\cite{music21} provides synchronized audio-visual recordings of 21 different musical instruments. This study utilizes only the subset corresponding to solo performances. During processing, all video clips were segmented into independent, non-overlapping one-second windows. This subset was randomly partitioned into training, validation, and test sets, containing 146,908, 7,103, and 42,440 samples, respectively.

\textbf{AVE.} The AVE~\cite{AVE} dataset consists of 4,143 video clips distributed across 28 event categories. In the processing pipeline, each clip was segmented into non-overlapping one-second windows, ensuring these windows were temporally aligned with the synchronized annotations. The final data split (train/validation/test) contains 27,726, 3,288, and 3,305 samples, respectively.

\subsubsection{Implementation Details}

We follow the processing pipeline used in previous distillation methods~\cite{DM,DM2} and adopt a unified convolutional neural network (ConvNet)~\cite{conv} architecture for both audio and visual inputs. Specifically, the audio processing branch consists of three basic blocks, each comprising a convolutional layer, a normalization layer, a ReLU activation function, and a pooling layer. For visual inputs with a larger spatial resolution (224 × 224), the network is extended to five blocks of the same type. During model training, we employ stochastic gradient descent with a momentum of 0.9 for both the pre-training and validation phases, and the learning rate is consistently set to 0.2. The initialization of synthetic data is achieved by selecting audio–visual paired samples using the Herding method~\cite{herd}, and the batch size is fixed at 128 during training. In particular, we incorporate the Factor technique~\cite{kim2022dataset,DM2} to enhance the representational capacity of the synthetic data, where the factor parameter $l$ is fixed to 2. The raw audio signals are recorded at a sampling rate of 11 kHz and subsequently converted into 128 × 56 log mel-spectrograms, in alignment with the AVDD~\cite{AVDD} setup. For more details on the hyperparameter settings, refer to Appendix B. 

We compare DAVDD against several representative audio-visual dataset distillation methods, including AVDD~\cite{AVDD}, DM~\cite{DM}, DC~\cite{DC}, DSA~\cite{DSA}, MTT~\cite{MTT}, as well as coreset selection methods, Random and Herding~\cite{herd}. A detailed introduction to all comparison methods is provided in Appendix C. All reported numbers represent the average performance over five independent runs. All our experiments were conducted on a single RTX A6000 Ada GPU. 

\begin{figure*}[t]
\centerline{\includegraphics[width=1\linewidth]{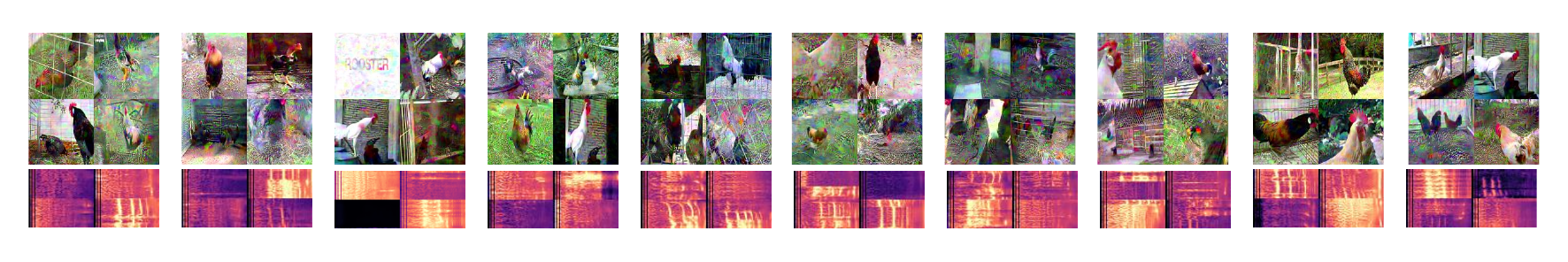}}
\caption{The figure shows the synthesized samples for the class ``chicken crowing" from the VGGS-10K dataset with IPC = 10. The top row shows the distilled visual data, while the bottom row shows the corresponding visualized spectrograms of the distilled audio.}
\label{s_dataset}
\end{figure*}

\subsection{Main Results}

As shown in Table~\ref{main}, we compare the performance of our method (DAVDD) against several methods on the VGGS-10K~\cite{vggsound}, MUSIC-21~\cite{music21}, and AVE~\cite{AVE} datasets. Across all three audiovisual datasets and under varying IPC settings, DAVDD consistently achieves superior performance, significantly outperforming other SOTA methods.

Compared to AVDD, DAVDD demonstrates stable improvements across all datasets. For instance, on VGGS-10K at IPC=1/10/20, DAVDD's performance is higher by 1.7\%, 2.2\%, and 1.4\%, respectively. The corresponding gains on MUSIC-21 are 1.5\%, 0.9\%, and 1.7\%, while on AVE, the improvements range from 0.9\% to 1.4\% percentage points. The advantage of DAVDD over DM is even more pronounced, particularly on MUSIC-21 at IPC=1, where its performance surpasses DM by over 7\%. This result indicates that the proposed method can more effectively distill the semantic correlations within multimodal audiovisual data. Furthermore, as IPC increases from 1 to 20, the performance of all methods exhibits a consistent upward trend. Notably, at high IPC settings, the performance of DAVDD already approaches the level achieved using the full training set, which demonstrates its effectiveness as a robust multimodal dataset distillation approach. Figure~\ref{s_dataset} shows the visualization of the distilled VGGS-10K dataset. For other distilled datasets, refer to Appendix D.

\subsection{Ablation Study}

Table \ref{tab:ablation} presents an effectiveness study conducted on various module combinations of DAVDD, where CIM stands for Common Intermodal Matching. It is important to note that the baseline refers to the original DM framework. Moreover, when the Common Intermodal Matching option is not selected, matching of shared representations occurs only within individual modalities. The experimental results demonstrate that the proposed method achieves superior performance. Compared with the original DM approach, randomly sampling pre-trained pairs from the pretrained bank leads to a performance improvement of 5.3\%. Furthermore, when the decoupler bank is incorporated, the performance is enhanced by 5.7\%, indicating that not only are private representations preserved, but the modality-specific associations between audio and visual modalities are also effectively distilled into the synthesized dataset. To further assess the generalization ability of the distilled data, we additionally conduct cross-architecture evaluations; details are provided in Appendix E.

\begin{table}[h]
\centering
\caption{Ablation study results on VGGS-10K with IPC = 10.}
\label{tab:ablation}
\scalebox{0.75}{
\begin{tabular}{cccc|c}
\toprule
\multicolumn{4}{c|}{Module} & \multicolumn{1}{c}{VGGS-10K} \\
\cmidrule(lr){1-4} 
Baseline & Pre-trained Bank & Decoupler Bank & CIM
& IPC 10 \\
\midrule
$\checkmark$ & --          & --    & --      & 43.9$\pm$1.8\\
$\checkmark$ &  $\checkmark$        & --  & --   & 49.2$\pm$1.8    \\
$\checkmark$ & $\checkmark$ & $\checkmark$       & --   & 54.9$\pm$1.1    \\
\rowcolor{gray!30} $\checkmark$ & $\checkmark$& $\checkmark$  & $\checkmark$   & \textbf{56.2$\pm$1.0}    \\
\bottomrule
\end{tabular}}
\end{table}

\begin{figure}[h]
\centerline{\includegraphics[width=0.8\linewidth]{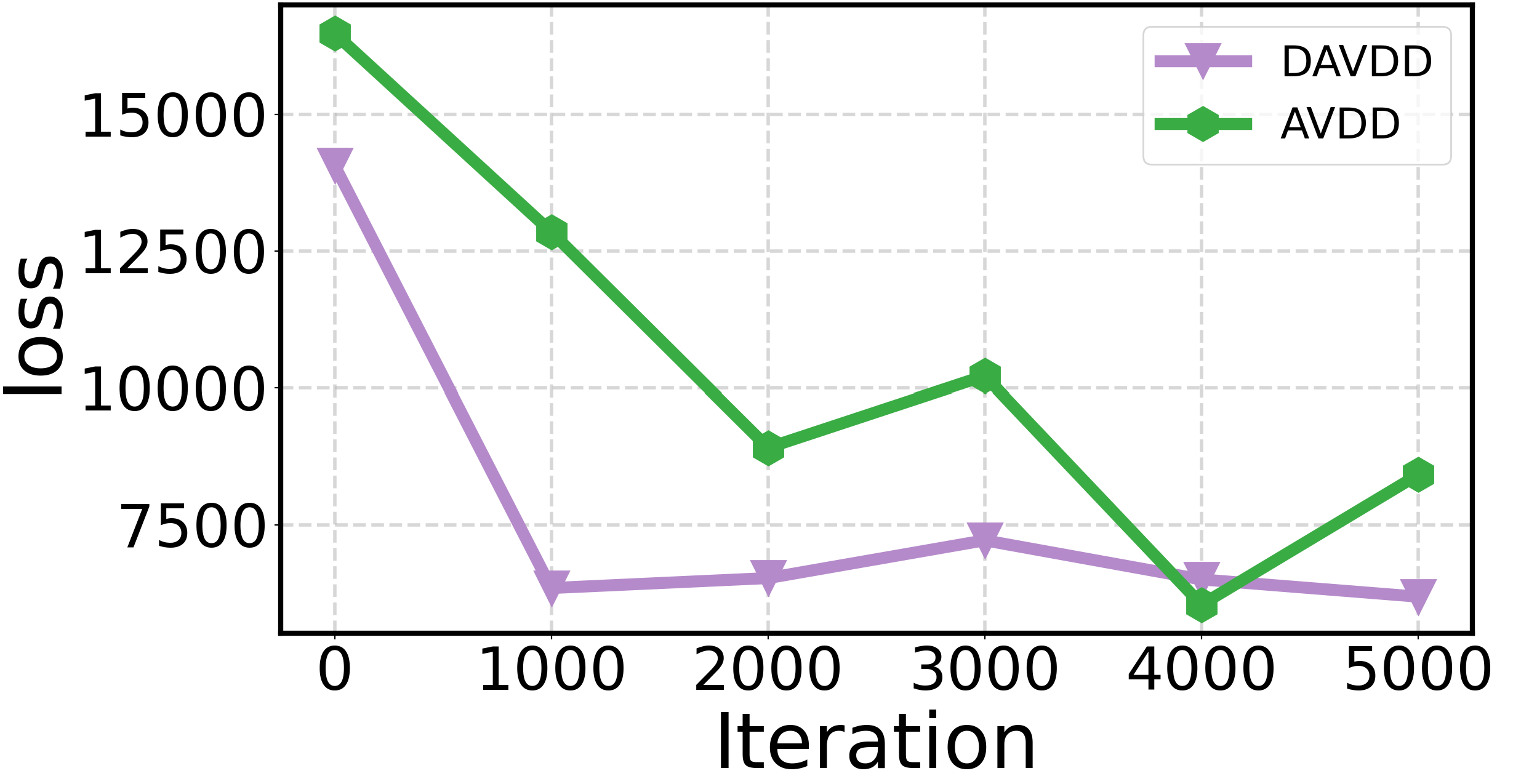}}
\caption{The losses in the AVDD and DAVDD during the distillation process.}
\label{loss}
\end{figure}

\subsection{Discussion}
\begin{figure}[h]
\centerline{\includegraphics[width=0.90\linewidth]{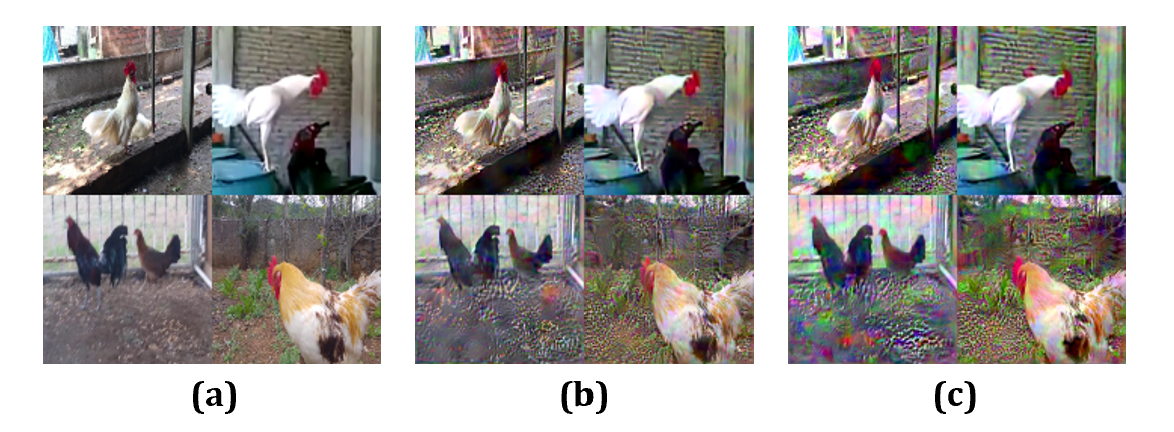}}
\caption{The figure presents the visualization of a sample image from the VGGS-10K dataset at IPC = 10 using different methods. (a) shows the original image, (b) shows the distilled image generated by the DM method, and (c) shows the distilled image generated by our DAVDD method.}
\label{s_img}
\end{figure}

As shown in Fig.~\ref{loss}, the loss fluctuations of DAVDD during the distillation process are significantly smaller than those of AVDD, demonstrating the robustness of our method.
As shown in Fig.~\ref{s_img}, we can observe that when there is no interaction between modalities (i.e., the DM method), the generated images are more similar to the original images, with only a small number of localized pixel perturbations. This indicates that the modality correlations were not integrated into the synthetic dataset during the distillation process. In contrast, the images distilled using the DAVDD method exhibit greater differences from the original images and contain more perturbations. This ensures that while the images still retain their visual features, the speech-related associations are also distilled effectively. Furthermore, the t-SNE embeddings of our synthetic dataset are provided in Appendix F.

\section{Conclusion}
In this work, we presented DAVDD, a pretraining-based and fully decoupled framework for audio–visual dataset distillation. By combining a diverse pre-trained bank with a lightweight decoupler bank, DAVDD cleanly separates modality-private and cross-modal shared representations, effectively mitigating the instability and interference found in prior audio–visual distillation methods. DAVDD preserves modality-specific information while enforcing stable cross-modal consistency at both sample and distribution levels. This design produces synthetic data with stronger semantic coherence and more reliable modality alignment. Extensive experiments across multiple benchmarks show that DAVDD achieves state-of-the-art performance under all IPC settings, highlighting the effectiveness of decoupled representation learning for scalable, high-quality audio–visual dataset distillation.

\newpage
{
    \small
    \bibliographystyle{ieeenat_fullname}
    \bibliography{main}
}
\clearpage
\setcounter{page}{1}
\maketitlesupplementary

\section{Implementation Details of the EMA Prototype Bank Update}
To provide a stable and globally consistent cross-modal alignment objective, we maintain a prototype memory bank $\mathcal{P}$ that contains the audio and visual centers of all categories. This memory bank does not participate in gradient back-propagation; instead, it is updated asynchronously after each parameter optimization step. Specifically, for any category $c$, we first compute the normalized mean feature $\mu_c$ within the current mini-batch, and then update the corresponding global prototype $P_c$ using a sample-count-based adaptive momentum strategy:
\begin{equation}
    P_c \leftarrow \mathcal{N}\bigl(m P_c + (1 - m)\mu_c\bigr),
\end{equation}
where $\mathcal{N}(\cdot)$ denotes $\ell_2$ normalization. The adaptive
momentum coefficient
\begin{equation}
    m = \frac{N_{\text{prev}}}{N_{\text{prev}} + N_{\text{curr}}}
\end{equation}
is dynamically determined by the historical cumulative number of samples
$N_{\text{prev}}$ and the number of samples $N_{\text{curr}}$ of this category in the current mini-batch. This update mechanism essentially computes the cumulative mean over all samples encountered during training, ensuring that the prototypes act as unbiased estimates of the category distribution centroids, thereby providing robust global anchors for the distributional constraints.

\section{Other hyperparameter settings}
For each dataset, we utilize 40 Pre-trained Pairs, with each pair equipped with four corresponding Decouplers. During the distillation process, all components are aligned within a 6,272-dimensional space. The specific configurations for different loss weights are detailed in Table~\ref{tab:loss_weights}.

\begin{table}[h]
    \centering
    \caption{Loss weights used in our experiments.}
    \label{tab:loss_weights}
    \begin{tabular}{lc}
        \toprule
        Loss weight & Value \\
        \midrule
        $\lambda_c$ &  40\\ 
        $\lambda_p$ &  80\\ 
        $\lambda_{com}$ &  2\\ 
        $\lambda_{fu}$ &  2\\ 
        $\lambda_{inter}$ &  1\\ 
        $\lambda_{intra}$ &  3\\ 
        $\lambda_{align}$ &  1\\ 
        \bottomrule
    \end{tabular}
\end{table}

\section{Details of Baseline Methods}
\textbf{Distribution Matching (DM)}~\cite{DM} pioneered the utilization of maximum mean discrepancy to align synthetic data distributions with real data.

{\setlength{\parindent}{0pt} \textbf{Dataset Condensation (DC)}~\cite{DC} addresses this by synthesizing a dataset designed to induce weight gradients that closely mimic those derived from the original large-scale dataset.

\textbf{Differentiable Siamese Augmentation (DSA)}~\cite{DSA} facilitates synthetic data learning by subjecting both real and synthetic data to identical random transformations, while ensuring the augmentations remain differentiable for gradient backpropagation.

\textbf{Matching Training Trajectories (MTT)}~\cite{MTT} matching the training trajectories induced by real and synthetic datasets.

\section{Other Distilled Images}

We present additional distilled images, as shown in Figures \ref{IPC1} and \ref{bus}, which correspond to the distilled images for IPC = 1 on the VGGS-10K dataset and for the “driving buses” class with IPC = 10, respectively.

\begin{figure}[h]    \centerline{\includegraphics[width=1\linewidth]{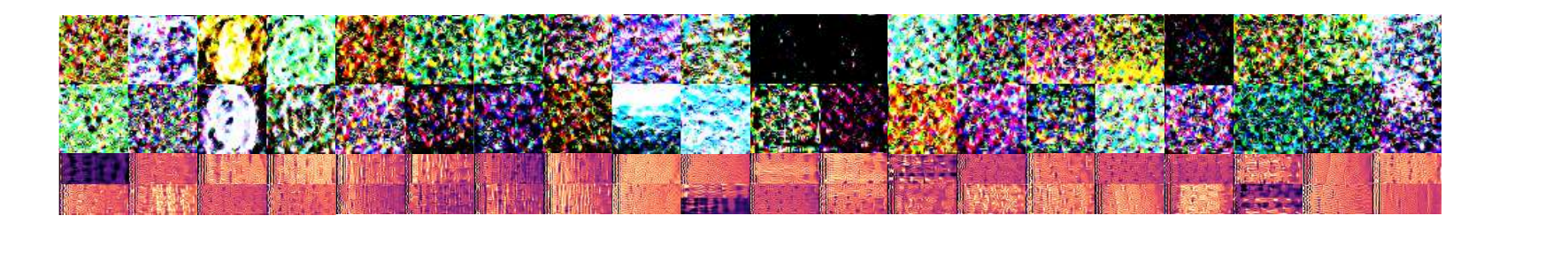}}
\caption{Distilled images on the VGGS-10K dataset with IPC = 1.}
\label{IPC1}
\end{figure}

\begin{figure}[h]    \centerline{\includegraphics[width=1\linewidth]{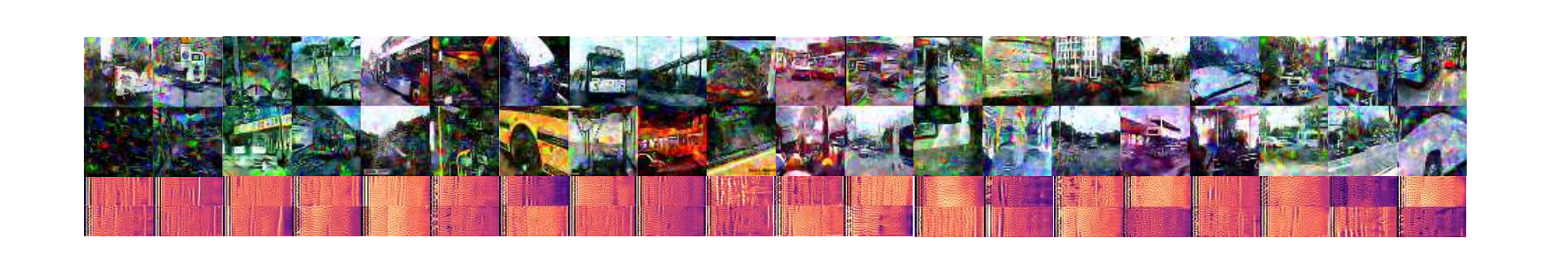}}
\caption{Distilled images for the “driving buses” class on the VGGS-10K dataset with IPC = 10.}
\label{bus}
\end{figure}

\section{Cross-Architecture Experiment}

Table~\ref{cross} reports the cross-architecture generalization performance on the VGGS-10K dataset with $\text{IPC}=1$. We evaluate the synthetic data distilled by various methods across five distinct architectures: ConvNet, VGG11, LeNet, ResNet18, and AlexNet. As demonstrated in the table, the proposed DAVDD consistently outperforms the baseline methods across the majority of evaluated networks. Specifically, DAVDD achieves state-of-the-art accuracy on four out of five architectures, securing top performances of $41.9\%$ on ConvNet and $31.0\%$ on ResNet18. Notably, while AVDD shows competitive performance on LeNet, DAVDD demonstrates significantly stronger generalization capabilities on deeper and more complex architectures, such as VGG11 and ResNet18, surpassing the second-best method by distinct margins.

\begin{table}[h]
\centering
\caption{The distillation accuracy of VGGS-10K (IPC = 1) for cross-architecture generalization.}
\label{cross}
\scalebox{0.7}{
\begin{tabular}{lccccc}
\toprule
Method & \multicolumn{5}{c}{Evaluation Architectures} \\
\cmidrule(lr){2-6}
& ConvNet & VGG11 & LeNet & ResNet18 & AlexNet \\
\midrule
MTT   & $34.1_{\pm 3.6}$ & $30.2_{\pm 1.3}$ & $23.9_{\pm 3.1}$ & $24.7_{\pm 1.5}$ & $22.6_{\pm 1.4}$ \\
DM    & $36.5_{\pm 2.5}$ & $27.0_{\pm 3.6}$ & $26.1_{\pm 2.7}$ & $22.0_{\pm 1.0}$ & $26.7_{\pm 1.8}$ \\
AVDD  & $40.2_{\pm 1.6}$ & $34.0_{\pm 1.8}$ &
         $\mathbf{32.4}_{\pm 5.2}$ & $30.5_{\pm 2.0}$ &
         $31.7_{\pm 2.7}$ \\
DAVDD   & $\mathbf{41.9}_{\pm 1.2}$ & $\mathbf{34.6}_{\pm 1.6}$ &
         $31.8_{\pm 3.7}$ & $\mathbf{31.0}_{\pm 1.6}$ &
         $\mathbf{31.9}_{\pm 1.9}$ \\
\bottomrule
\end{tabular}}
\end{table}

\begin{figure}[t]    \centerline{\includegraphics[width=1\linewidth]{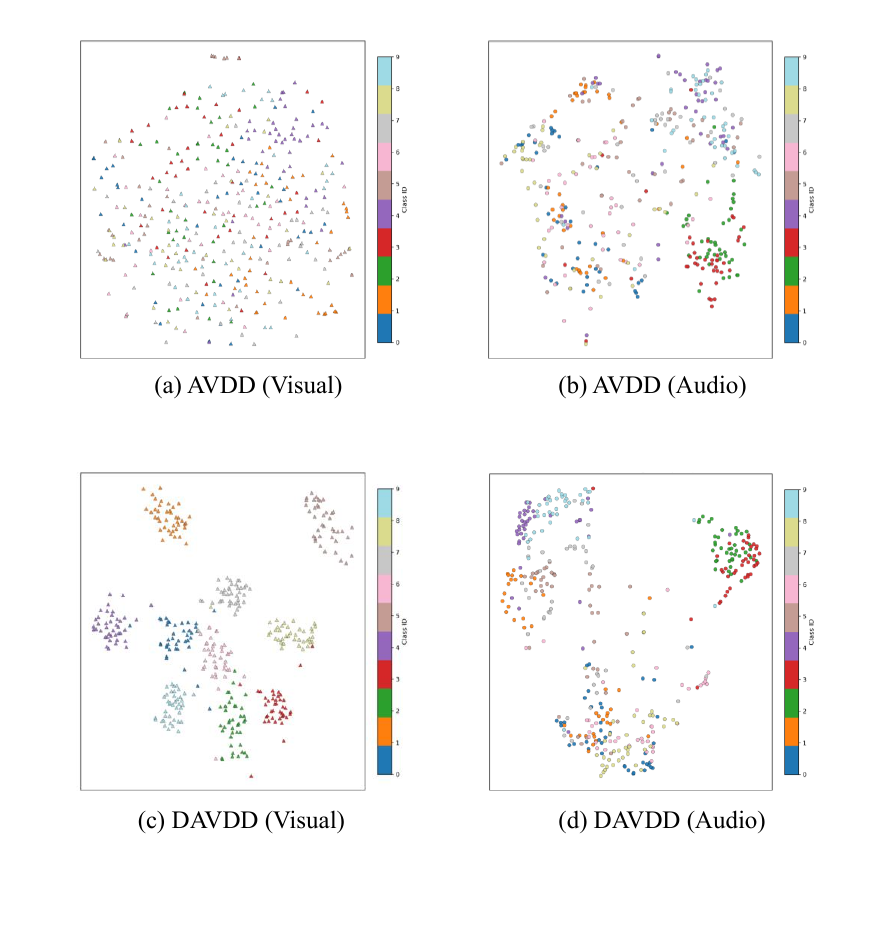}}
\caption{t-SNE visualization of the synthetic AVDD and DAVDD data.
(a), (b) show the visual and audio modalities of the AVDD synthetic data, respectively, while (c) and (d) depict the visual and audio modalities of the DAVDD synthetic data.}
\label{tsne}
\end{figure}

\section{t-SNE visualization}
We generate synthetic AVDD and DAVDD data based on VGGS-10K (IPC10), extract features using the same convolutional network as in the main experiments, and visualize their distributions using t-SNE, as shown in Fig.~\ref{tsne}. We observe that, in the visual modality, subfigure (c) exhibits more compact clusters and larger inter-class margins, substantially mitigating the class mixing observed in subfigure (a). This indicates that the DAVDD synthetic data in (c) achieves better visual feature discriminability and class separability than the AVDD synthetic data in (a). In contrast, the difference between subfigures (b) and (d) in the audio modality is less pronounced; nevertheless, the clusters in (d) appear slightly more compact with reduced inter-class overlap, suggesting a modest improvement in audio-wise class separability for DAVDD over AVDD.

\end{document}